\documentclass[10pt, a4paper, onecolumn]{article}

\usepackage[numbers]{natbib}

\usepackage[T1]{fontenc}
\usepackage{lmodern}

\usepackage{natbib}
\usepackage[utf8]{inputenc} 
\usepackage{booktabs}       
\usepackage{amsfonts}       
\usepackage{nicefrac}       
\usepackage{microtype}      
\usepackage{xcolor}         
\usepackage{graphicx}
\usepackage{amsmath,amssymb,amsfonts}
\usepackage{bm}
\usepackage{url}

\usepackage{array,multirow,graphicx}

\usepackage{geometry}
\newgeometry{vmargin={15mm}, hmargin={12mm,17mm}} 

\usepackage{algorithm}
\usepackage{algpseudocode}

\title{Area of interest adaption using feature importance}

\author{
	Wolfgang Fuhl
	\and
	Susanne Zabel
	\and
	Theresa Harbig
	\and
	Julia Astrid Moldt
	\and
	Teresa Festl Wiete
	\and
	Anne Herrmann Werner
	\and
	Kay Nieselt
	}

\date{}

\begin{document}
	
	\maketitle
	
	\begin{abstract}
		In this paper, we present two approaches and algorithms that adapt areas of interest (AOI) or regions of interest (ROI), respectively, to the eye tracking data quality and classification task. The first approach uses feature importance in a greedy way and grows or shrinks AOIs in all directions. The second approach is an extension of the first approach, which divides the AOIs into areas and calculates a direction of growth, i.e. a gradient. Both approaches improve the classification results considerably in the case of generalized AOIs, but can also be used for qualitative analysis. In qualitative analysis, the algorithms presented allow the AOIs to be adapted to the data, which means that errors and inaccuracies in eye tracking data can be better compensated for. A good application example is abstract art, where manual AOIs annotation is hardly possible, and data-driven approaches are mainly used for initial AOIs. \\
		Link: \url{https://es-cloud.cs.uni-tuebingen.de/d/8e2ab8c3fdd444e1a135/?p=%2FAOIGradient&mode=list}
	\end{abstract}

	\section{Introduction}
	In this work, we deal with the automated adaptation of regions of interest (ROI) and areas of interest (AOI). AOIs are areas in the stimulus which are marked by researchers to analyze eye tracking data. This can be an algorithmic analysis as for example in driver observation~\cite{vetturi2020use} but also a qualitative analysis as it is the case in art history~\cite{massaro2012art}. In these two application areas, one can already see the first challenge, since AOIs do not always have to be static over the entire time course. This is the case, for example, with driver observation. Here the driver moves his head and the usual AOIs at the speedometer and the center console move in the scene image of the eye tracker. Of course there are other challenges like the initial setting of the AOIs. Here, art history provides an excellent example with abstract art. Art historians would like to compare and analyze the gaze behavior of several people on the stimulus (abstract artwork) but how do you define the AOIs~\cite{mitrovic2020does,ROIGA2018} on such a stimulus. In the literature there are already several approaches like overlaying a grid~\cite{mihajlov2017eye,ooms2014study}, clustering the eye tracking data itself~\cite{drusch2014analysing}, gradient-based splitting of the eye tracking heatmap~\cite{fuhl2015arbitrarily} or using the salinecy value of the stimulus~\cite{fuhl2018automatic}. All of these methods have found their way into applications and research, and these areas are diverse. An increasingly important area is eye tracking analysis in medicine~\cite{harezlak2018application} and self-diagnostic systems~\cite{levinson2019self}, in research for cognitive load~\cite{palinko2010estimating}, human attention~\cite{shagass1976eye,FCDGR2020FUHL}, eye movement prediction for foveated rendering~\cite{meng2018kernel,fuhl2018simarxiv,ICMIW2019FuhlW1}, monitoring people in safety-critical applications (autonomous driving~\cite{lotz2018predicting}, piloting~\cite{brazil2017potential}, military~\cite{peissl2018eye}), for understanding human behavior such as in programming~\cite{obaidellah2018survey}, determining a person's expertise~\cite{gegenfurtner2011expertise}, in teaching~\cite{jarodzka2021eye}, and many more. Another challenge involves the eye tracking data itself, and this concerns the accuracy. From the point of view of the analyst of the data, it is easy to define clear AOIs depending on the stimulus, but the computed eye-tracking data do not necessarily lie in this region even though the subject is looking at this region. In this case, there are already solutions like moving the data~\cite{zhang2011mode}, smoothing the eye tracking signals~\cite{vadillo2015simple} but of course also the already mentioned calculation of the AOIs on the data itself.
	
	In this paper, we present a new method which adapts already defined AOIs to the eye tracking data. For this, we use the usual statistics computed together with the AOIs and the eye tracking data (fixation count, fixation duration, saccade velocity etc.) in combination with the feature importance. In addition, one or more target variables have to be selected for which the AOIs should be optimized. This would be e.g. classification of expertise, optimal diagnosis, identification of a person based on the data and much more. The statistics, themselves, can be freely chosen by the analyst and our algorithm works with any number of values. At each step, our algorithm trains an arbitrary machine learning procedure, calculates the feature importance, maps the feature importance to the AOIs, calculates a gradient with respect to an optimal change in the AOIs, and then adapts the AOIs. In addition, we tested our algorithm with several methods for automated AOI computation and in all experiments, the classification accuracy was significantly improved.

	\section{Related Work}
	In this section, we present related work from the literature. Since our work can be classified into different subfields, we decided to categorize it into two subcategories. Those categories are AOI generation and AOI based scan path classification.
	
	\subsection{AOI generation}
	\begin{figure}
		\centering
		\includegraphics[width=0.2\textwidth]{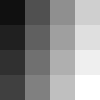}
		\includegraphics[width=0.2\textwidth]{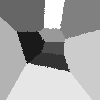}
		\includegraphics[width=0.2\textwidth]{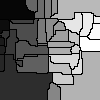}
		\caption{Exemplary AOIs computed as grid (left), based on gaze point clusters (central), and via the gradient segmentation on the heatmap (right). For the clustering based AOIs we assigned each unassigned pixel to the closes gaze point cluster.}
		\label{fig:aoigen}
	\end{figure}
	
	The simplest approach to define AOIs is the grid overlay~\cite{mihajlov2017eye,ooms2014study} (Figure~\ref{fig:aoigen} left image). Each cell in the grid represents one AOI, and each cell in the grid has usually the same size. A more data dependent approach is the usage of clustering methods like K-Means or Mean shift to define AOIs~\cite{drusch2014analysing,heminghous2006icomp,duchowski2010scanpath} (Figure~\ref{fig:aoigen} central image). In this approach, the raw eye tracking data or the eye movements like fixations are given to the clustering method. The computed clusters are in most cases data related and have to be extended to regions. This means that each data point has an assigned class, but the data is usually not spread over the entire stimulus. Therefore, the points of the same cluster have to be connected to form a region. This region is afterwards used as AOI. More advanced AOI generation algorithms use the segmentation of the eye tracking heatmap~\cite{fuhl2015arbitrarily,kubler2017subsmatch}. In \cite{fuhl2015arbitrarily} the gradient for each location on the heatmap is computed. Based on the gradient direction change, borders are defined on the heatmap. With those borders, the heatmap can be segmented into regions, which are the AOIs. This approach is similar to the watershed transformation. An alternative to the gradient based heatmap segmentation is the usage of confidence intervals with grid cells and the heatmap~\cite{kubler2017subsmatch}. In this approach, a grid is overlaid over the heatmap. For each cell in the grid, a normal distribution is fitted to the heatmap data. Based on the confidence interval of the fitted normal distributions, the grid cell size is adapted. Cells without any data in it are removed.
	
	Data independent methods are also present in the literature. The traditional way in this domain is the manual definition of AOIs based on objects in the stimulus~\cite{rosenberg2015moving}. Since manual annotation is a time-consuming process and not always possible, because the stimulus does not contain any object for example (abstract artwork), other approaches compute the AOIs directly on the stimulus. One possibility is the usage of the saliency value~\cite{fuhl2018automatic} as heatmap and use the gradient based AOI computation~\cite{fuhl2015arbitrarily}. The idea behind this approach is to use bottom up features of the stimulus to define the importance of regions~\cite{itti1998model}. In case of a dynamic stimulus (Driving, head mounted eye tracker, etc.) it is usually impossible to annotate the amount of generated data in a reasonable time. Therefore, object detection, segmentation, and identification methods have been utilized to compute AOIs in dynamic stimuli~\cite{jayawardena2021automated,WF042019,lagmay2022enhanced}. Those approaches are based on deep neural networks (DNN) for object detection, segmentation and classification~\cite{lecun2015deep}. 
	
	\subsection{AOI based scan path classification}
	AOI based scan path classification can be separated in two fields. The first uses the sequence of AOIs which are viewed by the subjects, and the second one is using statistics on the eye movements~\cite{ICMIW2019FuhlW2,EPIC2018FuhlW} on and between the AOIs~\cite{eraslan2016eye,sargezeh2019gender,kang2020identification,moore2018geometric,jiang2021correlation}. In case of the sequence of AOIs, the problem can be understood as string alignment, since we can assign each AOI a letter and compare afterwards those strings. In string alignment, there are two major algorithms which have to be mentioned here too. The first algorithm is the Levensthein distance which computes the minimal insertions of letters, and the second algorithm is Needleman Wunsch which uses dynamic programming to compute the optimal alignment. In ScanMatch~\cite{cristino2010scanmatch}, the Needleman Wunsch algorithm is used on AOI sequences to compute the minimal change to convert one scan path to the other. They also found a solution to incorporate the fixation duration by repeating the letter. An alternative to the AOI letter assignment was proposed in \cite{zangemeister1996evidence}. In this work, the letters are assigned based on the saccades between AOIs and not the AOIs itself. In addition, each connection between AOIs received different weighting. The first totally data driven AOI approaches is proposed in \cite{privitera2000algorithms} where the AOIs are computed on clusters, afterwards letters are assigned to fixations, which is followed by string alignment. A visual tool based on this technique (iComp~\cite{heminghous2006icomp, duchowski2010scanpath}) was also proposed. Hidden Markov Models (HMM) were also used to compare scan path~\cite{kanan2014predicting,engbert2001mathematical}. Here the AOIs are modeled as states in the HMM and the transitions between the AOIs are the transition probabilities of the HMM. SubsMatch~\cite{kubler2017subsmatch} is a scan path comparison algorithm which proposed a novel feature which computes the triplet distribution. This triplet distribution is based on the AOI sequence, where each triplet occurrence has one bin in the distribution. Afterwards, this distribution can be used with any machine learning approach for classification or regression. A faster approach is proposed in \cite{geisler2020minhash}. Here the authors use larger word sizes than triplets and store them in a dictionary. Since the dictionary size and therefore the comparison costs grow exponentially with increased word size, the authors proposed to use the min hash approach here. This means they only compare a subset of word frequencies, instead of all.
	
	There exist also scan path classification algorithms which do not rely on AOIs. For completeness, we give a short summery. iMap~\cite{caldara2011imap,lao2015imap} is a scan path comparison algorithm based on three-dimensional fixation maps (x, y, fixation density). Two fixation maps are compared pixel per pixel. An alternative to the heat map (3D fixation map) comparison is an approach to map the fixations of two scan path to each other~\cite{mannan1996relationship}. Since gaze recordings usually have different length, an extension was proposed which allows to map multiple fixations of one scan path to one fixation of the other scan path~\cite{mathot2012simple}. Another geometrical approach is Multimatch~\cite{jarodzka2010vector,dewhurst2012depends}. In the first step, the scan path is simplified by deleting and merging small saccades. Afterwards, the scan path is converted into a vector with multiple dimensions like location and duration. The mapping of fixations between two scan path is done using the Dijkstra algorithm. An algorithm using only saccade directions is proposed in \cite{fuhl2019ferns,fuhl2022hpcgen}. It selects different angles with allowed ranges and stores them in ferns (A type of random forest classifier). Sub scan path classification is a way to overcome the challenge of different length in the recordings and was proposed in \cite{foerster2013functionally}. Here the statistics like fixation duration etc. are computed on the sub scan path and later used for comparison. Approaches with to model the input data for deep neural networks were also proposed~\cite{fuhl2019encodji}.

	\section{Method}
	In this section, we describe the two proposed algorithms for AOI adaption. The first section describes the greedy direct approach, which grows in all directions equally. In the second subsection, the gradient based approach is described, which is derived from the direct approach. For the gradient based approach, each AOI is subdivided to compute an additional gradient and also use the initial feature importance as gradient strength.
	
	\subsection{Direct approach}
	\label{sec:direct}
	\begin{algorithm}
		\label{alg:direct}
		\caption{The algorithm for the direct computation of the new AOIs. In each iteration we compute a new train and validation split and only allow equal or improved results to pass into the next iteration. The AOI growing is performed pixel wise and described in Equation~\ref{eq:growsimple}.}
		\begin{algorithmic}
			\Require EyeTrackingDataTrain, AOIs, ML-Method, Statistics, EvalFunc, Iterations
			\For{$i \leq Iterations$}
			\State [TrainSet, ValidationSet] = RandomSplit(EyeTrackingDataTrain);
			\State TrainFeatures = ComputeFeatures(TrainSet,AOIs,Statistics);
			\State Model = Train(TrainFeatures,MLMethod);
			\State ValidationFeatures = ComputeFeatures(ValidationSet,AOIs,Statistics);
			\State Metric=EvalFunc(Model,ValidationFeatures);
			\State FI=FeatureImpotance(Model);
			\State FIAOI=AggregateFeatureImpotanceToAOI(FI,AOIs); \Comment{\underline{\textit{Sum over the statistics per AOI}}}
			\State AOInew=GrowAOIs(FIAOI,AOIs); \Comment{\underline{\textit{Overwrite smaller feature importances in all directions}}}
			\State TrainFeatures = ComputeFeatures(TrainSet,AOInew,Statistics);
			\State Model = Train(TrainFeatures,MLMethod);
			\State ValidationFeatures = ComputeFeatures(ValidationSet,AOInew,Statistics);
			\State MetricNew=EvalFunc(Model,ValidationFeatures);
			\If{$MetricNew \geq Metric$}
			\State AOIs=AOInew;
			\EndIf
			\EndFor
		\end{algorithmic}
	\end{algorithm}
	
	The direct approach is described algorithmically in Algorithm~\cite{eq:growsimple}. In the first step, a train and validation split is performed. This split is used to evaluate the new AOIs and discard them if they worsen the result. In addition, this split is repeated in each iteration to avoid bad splits having a large impact and to increase the variation in the feature importance values. Afterwards, a model is trained and evaluated on the validation set. Based on this model, the feature importance is computed and aggregated to each AOI. This aggregation is simply summing up the features, which are computed on an AOI (The statistical values). With the aggregated feature importance values, we start the AOI adaption, which is described in Equation~\ref{eq:growsimple}.
	
	\begin{equation}
		\label{eq:growsimple}
		\small
		GrowAOIs(FIAOI,AOIs) = \\
		\begin{cases}
			AOIs(x,y) & \bold{if}~FIAOI(AOIs(x',y')) < FIAOI(AOIs(x,y)) \\
			AOIs(x',y') & \bold{else}
		\end{cases}
	\end{equation}
	Equation~\ref{eq:growsimple} describes the growing procedure of the direct approach. $FIAOI$ is the feature importance per AOI ($AOIs$ is here assumed to be in 2D space with the same size as the stimulus but our algorithm can also use the time, so it would be a 3D space). The position $x,y$ checks its direct neighbors ($x',y'$) and overwrites them if they have a lower feature importance. This way, each AOI can only grow one pixel in each direction.
	
	After the new AOIs were computed, we train a new model and evaluate the new AOIs. If they perform better in comparison to the last ones under a given metric, we overwrite the old AOIs and start the next iteration.
	
	\subsection{Gradient approach}
	\begin{figure}
		\centering
		\includegraphics[width=0.5\textwidth]{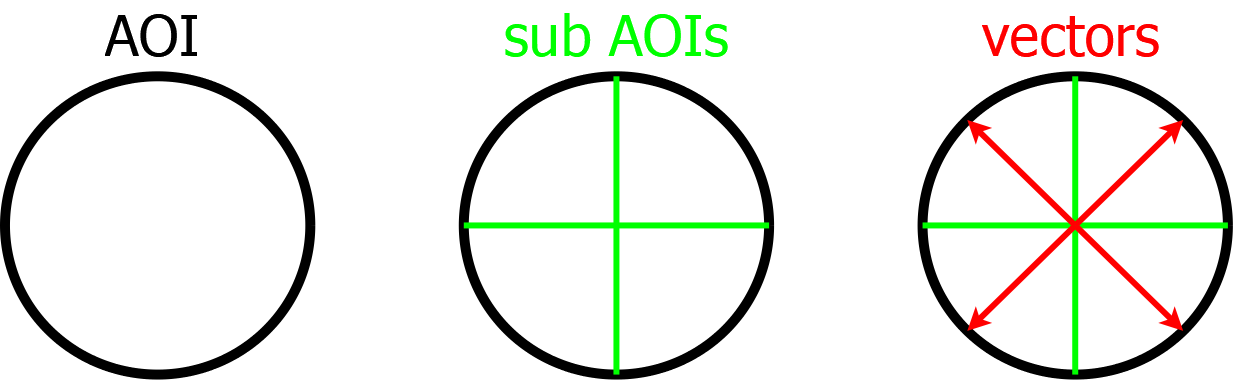}
		\caption{Description of an exemplary sub AOI and vector computation. The left image shows the initial AOI which is split into sub AOIs (Green borders in the central image). Afterwards, the vectors of each sub AOI are computed (Red arrows in the right image).}
		\label{fig:subAOIs}
	\end{figure}
	
	\begin{algorithm}
		\label{alg:gradient}
		\caption{The algorithm for the gradient based computation of the new AOIs. In each iteration we compute a new train and validation split and only allow equal or improved results to pass into the next iteration as it is done for the direct approach. The AOI growing is performed based on the feature importance of the current AOI and described in Equation~\ref{eq:growgrad}. For the gradient computation itself we use Equation~\ref{eq:gradcomp} and the sub AOI split can be seen in Figure~\ref{fig:subAOIs}.}
		\begin{algorithmic}
			\Require EyeTrackingDataTrain, AOIs, AOISplitFunc, ML-Method, Statistics, EvalFunc, Iterations
			\For{$i \leq Iterations$}
			\State [TrainSet, ValidationSet] = RandomSplit(EyeTrackingDataTrain);
			\State TrainFeatures = ComputeFeatures(TrainSet,AOIs,Statistics);
			\State Model = Train(TrainFeatures,MLMethod);
			\State ValidationFeatures = ComputeFeatures(ValidationSet,AOIs,Statistics);
			\State Metric=EvalFunc(Model,ValidationFeatures);
			\State FI=FeatureImpotance(Model);
			\State FIAOI=AggregateFeatureImpotanceToAOI(FI,AOIs); \Comment{\underline{\textit{Sum over the statistics per AOI}}}
			\Comment{\underline{\textit{FIAOI is the gradient strength or growing length}}}
			\State [SubAOIs, Vectors]=SplitAOIs(AOIs,AOISplitFunc);
			\State TrainFeatures = ComputeFeatures(TrainSet,SubAOIs,Statistics);
			\State Model = Train(TrainFeatures,MLMethod);
			\State FI=FeatureImpotance(Model);
			\State FISubAOIs=AggregateFeatureImpotanceToAOI(FI,SubAOIs); \Comment{\underline{\textit{Sum over the statistics per AOI}}}
			\State SubAOIsGradient=ComputeGradientDirection(FISubAOIs,Vectors);
			\State AOInew=GrowAOIsGrad(SubAOIsGradient,FIAOI,AOIs); \Comment{\underline{\textit{Overwrite in gradient direction}}}
			\State TrainFeatures = ComputeFeatures(TrainSet,AOInew,Statistics);
			\State Model = Train(TrainFeatures,MLMethod);
			\State ValidationFeatures = ComputeFeatures(ValidationSet,AOInew,Statistics);
			\State MetricNew=EvalFunc(Model,ValidationFeatures);
			\If{$MetricNew \geq Metric$}
			\State AOIs=AOInew;
			\EndIf
			\EndFor
		\end{algorithmic}
	\end{algorithm}
	
	The gradient based AOI adaption is based on the direct approach described in Section~\ref{sec:direct} and algorithmically described in Algorithm~\ref{alg:gradient}. It is equal to the direct approach up to the point of the first feature importance computation. After this step, the AOIs are divided in sub AOIs as can be seen in Figure~\ref{fig:subAOIs}. We get the sub AOIs as well as vectors of their contribution to the final gradient. With the sub AOIs we train a new model and compute the feature importance per AOI (Again summing up the feature importance values per sub AOI over all statistical values). With those feature importance values of the sub AOIs we compute the gradient as described in Equation~\ref{eq:gradcomp}.
	
	\begin{equation}
		\label{eq:gradcomp}
		\small
		ComputeGradientDirection(FISubAOIs,Vectors) = |\sum_{j=1}^{n} FISubAOIs(i,j)* Vectors(i,j)|
	\end{equation}
	Equation~\ref{eq:gradcomp} describes the gradient computation. Here $||$ is the euclidean norm, $i$ the global AOI index, $j$ the sub AOI index, $FISubAOIs$ the feature importance per sub AOI, and $Vectors$ the vector in the direction of the sub AOI from the central point of the global AOI the sub AOI belongs to.
	
	With the gradient and the feature importance of the global AOIs, which we use as gradient strength, we can adapt the AOIs. We use the gradient direction to find possible AOI locations which will be overwritten if the global feature importance is lower in comparison to the feature importance value of the outgoing AOI. The distance of those locations can be up to the feature importance value of the current AOI. This step is described mathematically in Equation~\ref{eq:growgrad}.
	
	\begin{equation}
		\label{eq:growgrad}
		\small
		\begin{aligned}
			& GrowAOIsGrad(SubAOIsGradient,FIAOI,AOIs) = \\
			& AOIs(x,y)~\bold{if}~FIAOI(AOIs((x,y)+(SubAOIsGradient*FIAOI(AOIs(x,y))))) < FIAOI(AOIs(x,y)) \\
			& AOIs((x,y)+(SubAOIsGradient*FIAOI(AOIs(x,y))))~\bold{else}
		\end{aligned}
	\end{equation}
	Equation~\ref{eq:growgrad} describes the gradient based growing proceadure. $SubAOIsGradient$ are the gradient directions as vectors, $FIAOI$ the feature importance per AOI, and $AOIs$ the multidimensional map with AOI indexes (Here assuemd to be two dimensional). For each AOI it is check if there exists a pixel of another AOI in gradient direction ($SubAOIsGradient$) with a lower feature importance. The distance of this pixel can be up to the feature importance of the current AOI ($FIAOI(AOIs(x,y))$). If this is not the case, nothing is change as stated in the $else$ part.
	
	With the new AOIs, we train a model again to evaluate them based on the given metric. If the new AOIs perform better or equal, we overwrite the old AOIs and start a new iteration as in the direct approach. 
	
	\section{Evaluation}
	\label{sec:eval}
	
	\begin{figure}
		\centering
		\includegraphics[width=0.2\textwidth]{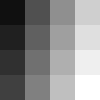}
		\includegraphics[width=0.2\textwidth]{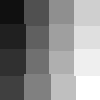}
		\includegraphics[width=0.2\textwidth]{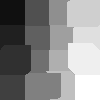}
		\caption{The initial grid based AOIs are shown on the left. In the central image the adapted AOIS with the direct approach for the WM data set are shown. The image on the right shows the resulting AOIs of the gradient based approach for the WM data set.}
		\label{fig:finalAOIs}
	\end{figure}

	For testing and evaluating our algorithm, we have chosen two public datasets and one which will be published together with this paper. In the following, we briefly describe each dataset and refer to the corresponding paper for a more detailed description.
	
	\textbf{Website Mouse~\cite{bscthesis,GMS2021FUHL} (WM):} This data set was collected from eleven subjects in the age of 20 to 38. Each subject had to fulfill five tasks namely read an article, online shopping, search on Google Maps, search on Google, and online games. In all four tasks the subject had to fulfill three subtasks like buy everything for a meal or search for specific information etc. In the background, we recorded the mouse movements while the subject used his browser to fulfill the task. Overall we had 165 recordings of the subjects and for each task the participants had as much time as they wanted. Since the mouse movements and the gaze location correlates, we think this is a good additional data set for eye tracking evaluations.
	
	\textbf{ETRA Challenge~\cite{mccamy2014highly} (ETRAC):} This data set consists of 960 trials with a recording length of approximately 45 seconds. The performed tasks are visual fixation, visual exploration, and visual search. Each subject participated in three sessions with a session duration of approximately 60 minutes. In total, they recorded eight subjects whereby six are females and two males. The eye tracker was an EyeLink II from SR Research in Canada. For high quality recordings, the authors also used a chin rest and placed the participant approximately 57 centimeters away from the screen. The used screen was a Barco Reference Calibrator V with a refresh rate of 75 Hz and a display size of $40 \times 30$ centimeters.
	
	\textbf{Hollywood~\cite{costela2019free} (HOLLY):} For this data set the researchers used an Eyelink 1000 eye tracker to record 95 subjects watching hollywood video clips. They provided 217 video clips in total whereby 206 had 30 seconds duration and 11 a duration of 30 minutes. The 30 seconds video clips were viewd by 76 participants and the 30 minutes video clips by 19. Fr each participant they stored demographic information like gender, age, ethnicity, and education. For each clip the subject rated in eight cathegories (genre, importance of human faces, importance of human figures, importance of man-made objects, importance of nature, auditory information, lighting, and environment type). We could only download the data of 63 subjects for the short video clips, and therefore we only used those.
	
	For feature importance computation, we used the out-of-bag estimates from Matlab 2022a together with the bagged decision tree ensemble of Matlab 2022a. We used the default parameters of Matlab to simplify the reproduce ability of our results. For the initial AOIs we used 75\% as training and 25\% as testing data. The split was done based on the different stimulus or tasks so that the same stimulus is not present in the training and testing data. We selected the stimulus as separation criteria, since our target classification variable was the subject ID. For the AOI adaption approaches we split the training data so that we had 50\% for training, 25\% for validation and the 25\% for testing. The testing data was for all approaches the same. As statistical features we computed the mean, median, mode, standard deviation, and skewness of the x and y coordinate for all gaze points in an AOI. Additionally, we used the amount of gaze points per AOI. In total, these are eleven features per AOI. The gradient and the direct approach for AOI adaption run exactly 50 iterations. The evaluation PC has a Windows 10 operating system, 64 GB DDR4 Ram, and an AMD Ryzen 3950X processor with 16 cores and 3.5 GHz.

	\begin{table}
		\centering
		\caption{The results of our approaches (Under Opti) compared to the initial AOIs (Under Init). The used metric is the average accuracy over all classes, which means that each class contributes equally to the accuracy. We used the subject ID as target class since it is much more difficult compared to the stimulus classification. As can be seen, the adapted AOIs always outperform the initial AOIs, although the ML method had less training data. Best results in bold.}
		\label{tbl:resultsDirectGradient}
		\begin{tabular}{c|c|c|cc|cc|cc}
			AOI  & AOI  & Parameter & \multicolumn{6}{c}{Data set} \\ 
			adaption& method &  & \multicolumn{2}{c}{WM} & \multicolumn{2}{c}{ETRAC} & \multicolumn{2}{c}{HOLLY} \\ 
			method& &  & Init & Opti & Init & Opti & Init & Opti \\ \cline{1-9}
			\multirow{7}{*}{\rotatebox{90}{Direct}} & K-Means & 5 & 57.95 & 64.28 & 36.25 & 43.33 & 28.71 & 32.37 \\
			& K-Means & 10 & 63.41 & 75.48 & 36.08 & 41.25 & 32.30 & 35.61 \\
			& K-Means & 20 & 70.73 & 78.73 & 37.50 & 40.83 & 35.37 & 38.69 \\
			& Grid & $4 \times 4$ & 72.72 & 82.14 & 42.08 & 44.16 & 23.84 & 26.21 \\
			& Grid & $5 \times 5$ & 69.15 & 74.18 & 39.16 & 41.66 & 18.27 & 22.19 \\
			& Grid & $10 \times 10$ & 72.40 & 79.70 & 41.66 & 47.08 & 17.43 & 20.74 \\
			& Gradient & & 67.07 & 82.30 & 35.00 & 37.91 & 26.06 & 46.78 \\ \cline{2-9}
			\multirow{7}{*}{\rotatebox{90}{Gradient}} & K-Means & 5 &  57.95 & 64.28 & 36.25 & 45.25 & 28.71 & 34.32 \\
			& K-Means & 10 & 63.41 & 76.78 & 36.08 & 42.91 & 32.30 & 36.45 \\
			& K-Means & 20 & 70.73 & 79.18 & 37.50 & 40.83 & 35.37 & 41.13 \\
			& Grid & $4 \times 4$ & 72.72 & 82.14 & 42.08 & 45.83 & 23.84 & 27.10 \\
			& Grid & $5 \times 5$ & 69.15 & 75.32 & 39.16 & 42.91 & 18.27 & 23.94 \\
			& Grid & $10 \times 10$ & 72.40 & 81.33 & 41.66 & \textbf{48.41} & 17.43 & 21.23 \\
			& Gradient & & 67.07 & \textbf{83.80} & 35.00 & 39.57 & 26.06 & \textbf{49.15} \\ \cline{2-9}
			& \textit{Chance} &  & \textit{9.09} & \textit{9.09} & \textit{12.5} & \textit{12.5} & \textit{1.58} & \textit{1.58}
		\end{tabular}
	\end{table}
	
	In Table~\ref{tbl:resultsDirectGradient} the results as average accuracy (Each class weighted equally) of our approaches (Opti) compared with the initial (Init) AOIs is shown. We evaluated our approaches on three data sets and with three different AOI computation methods. For those with parameters (Grid and K-Means) we evaluated different parameters too. As can be seen in Table~\ref{tbl:resultsDirectGradient} the optimized AOIs always outperform the initial AOIs. In addition, the gradient approach performs either equally good or outperforms the direct approach. If we take a look at the best results (Bold in Table~\ref{tbl:resultsDirectGradient}), we can see that for the WM data set the improvement is more than 16\%, for the ETRAC data set more than 6\%, and for the HOLLY data set more than 23\%.

	\section{Limitations}
	The presented algorithm also has limitations, which are described here. On the one hand, this is the necessity of initial AOIs. These have the disadvantage that they already supply restrictions regarding the optimal classification result, this can be seen in the section evaluation~\ref{sec:eval}, since here the result consists of different AOIs and these do not reach all the same accuracy. Another disadvantage is that no new AOIs are created, only existing ones are modified. AOIs without relevance can be overwritten, but the algorithm itself does not provide the possibility to add new AOIs. This makes it impossible to find the absolute optimal AOIs, because it is uncertain how many are optimal.
	
	\section{Conclusion}
	In this paper, we showed an approach to optimize AOIs based on the feature importance. In all cases, the optimized AOIs improved the classification accuracy with less training data compared to the initial AOIs without adaption. This was shown on two public data sets and additionally on a novel data set with mouse recordings, which will be made publicly available together with this work. We evaluated three different types of initial AOI creation, whereby one was data independent (Grid) and two data dependent approaches (Clustering and gradient based segmentation).

	\bibliographystyle{plain}
	\bibliography{template}

\end{document}